\documentclass[11pt]{article}

\PassOptionsToPackage{table,xcdraw}{xcolor}

\newcommand\samethanks[1][\value{footnote}]{\footnotemark[#1]}

\usepackage[final]{acl}

\usepackage{times}
\usepackage{latexsym}

\usepackage[T1]{fontenc}

\usepackage[utf8]{inputenc}

\usepackage{microtype}

\usepackage{inconsolata}

\usepackage{graphicx}

\usepackage{xspace}
\usepackage{amsmath}
\usepackage{cleveref}
\usepackage{booktabs}
\usepackage{multirow}
\usepackage{pifont}
\usepackage[most]{tcolorbox}
\usepackage{enumitem}
\usepackage{amssymb}
\usepackage{makecell}
\usepackage[table,xcdraw]{xcolor}

\definecolor{myLightBlue}{HTML}{FFFFFF} %
\definecolor{myDarkBlue}{HTML}{2271B5} %
\definecolor{myDarkDarkBlue}{HTML}{264F73} %

\newif\ifdraft
\draftfalse

\definecolor{customgreen}{HTML}{04BC7D}
\definecolor{customred}{HTML}{EF3333}

\newcommand{\xmark}{\textcolor{customred}{\ding{55}}}

\newcommand{\benchmarkname}{\textsc{GuideDog}\xspace}
\newcommand{\benchmarkQAname}{\textsc{GuideDogQA}\xspace}

\newcommand{\SOne}{\textit{S1}\xspace}
\newcommand{\STwo}{\textit{S2}\xspace}
\newcommand{\SThree}{\textit{S3}\xspace}

\title{GuideDog: A Real-World Egocentric Multimodal Dataset\\
for Blind and Low-Vision Accessibility-Aware Guidance}

\author{
  {\bf Junhyeok Kim$^{\clubsuit}$}\thanks{~Equal Contribution.} \quad
  {\bf Jaewoo Park$^{\clubsuit}$}\samethanks \quad
  {\bf Junhee Park$^{\clubsuit}$} \quad
  {\bf Sangeyl Lee$^{\clubsuit}$} \\
  {\bf Jiwan Chung$^{\clubsuit}$} \quad
  {\bf Jisung Kim$^{\diamondsuit}$}\thanks{Work done while at SK Telecom.} \quad
  {\bf Ji Hoon Joung$^{\heartsuit}$}\samethanks \quad
  {\bf Youngjae Yu$^{\spadesuit}$} \\
  $^{\clubsuit}$Yonsei University \quad
  $^{\diamondsuit}$LG AI Research \quad
  $^{\heartsuit}$Euler Robotics \quad
  $^{\spadesuit}$Seoul National University \\
  \texttt{junhyeok@yonsei.ac.kr} \\
  \url{https://jun297.github.io/GuideDog/}
}

\begin{document}
\maketitle
\begin{abstract}

For people affected by blindness and low vision (BLV), safe and independent navigation remains a major challenge, impacting over 2.2 billion individuals worldwide. Although multimodal large language models (MLLMs) offer new opportunities for assistive navigation, progress has been limited by the scarcity of accessibility-aware datasets, because creating them requires labor-intensive expert annotation.

To this end, we introduce \benchmarkname, a novel dataset containing 22K image-description pairs (2K human-verified) capturing real-world pedestrian scenes across 46 countries. Our human-AI pipeline shifts annotation from generation to verification, grounded in established BLV guidance standards from experts and research, improving scalability while maintaining quality. We also present \benchmarkQAname, an 818-sample benchmark evaluating object recognition and depth perception. Experiments reveal that depth perception and adherence to these standards remain challenging for current MLLMs.

\end{abstract}

\section{Introduction}
\label{sec:introduction}

\begin{figure}[!t]
\includegraphics[width=0.48\textwidth]{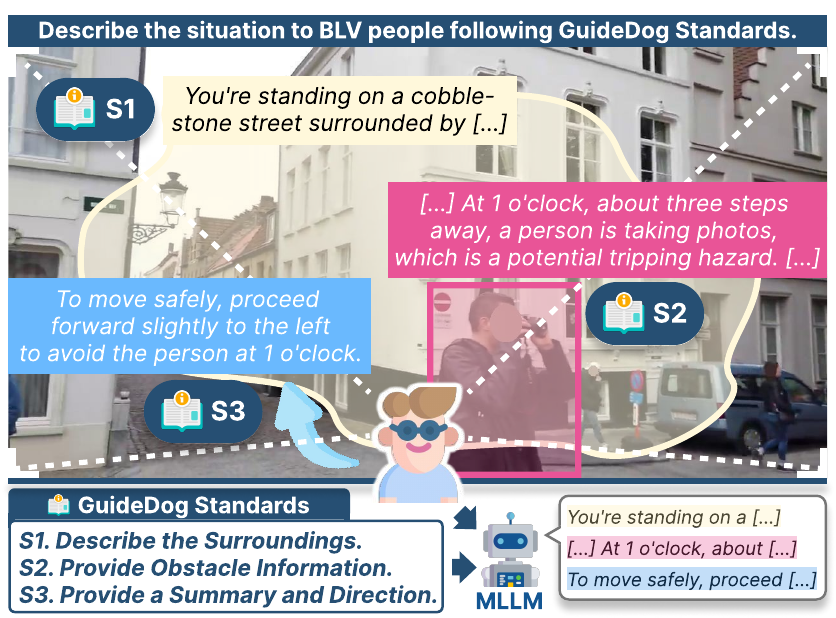}
\caption{An overview of the accessibility-aware guidance generation task, wherein MLLMs describe the overall situation (\SOne), identify hazardous obstacles (\STwo), and summarize recommended actions for BLV. (\SThree)}
\label{fig:standards_fig}
\end{figure}

An estimated 2.2 billion individuals worldwide are affected by blindness and low vision (BLV), including approximately 36 million who are completely blind~\cite{bourne2017magnitude}. For this population, safe and independent mobility remains a significant daily challenge, as navigating unfamiliar or obstacle-filled environments poses substantial risks. Prior work has highlighted the severity of this issue, reporting that approximately 7\% of visually impaired individuals experience falls at least once a month~\cite{manduchi2011mobility}. These challenges motivate the development of assistive technologies that support safe navigation and environmental understanding.

\begin{table*}[t!]
\centering
\resizebox{1.0\linewidth}{!}{
\begin{tabular}{l|cccccc|c}
\toprule
Dataset & \# Samples & Modality & Source & Geo-Diverse & Annotation & BLV Involvement & Task \\
\midrule
VizWiz~\citep{gurari2018vizwiz}        & 31K        & Image & User photos     & \xmark              & Human       & Captured by BLV         & VQA \\
VIALM~\citep{zhao2024vialm}            & 200        & Image & Web images      & \xmark              & Human       & Expert manual           & Guidance \\
\citet{merchant2024generating}         & 48         & Image & VizWiz          & \xmark              & Human       & \xmark                  & Guidance \\
WalkVLM~\citep{yuan2024walkvlm}        & 12K        & Video & Web + recorded  & 10 locations   & Human       & Survey                  & Guidance \\
EgoBlind~\citep{xiao2025egoblind}      & 1.3K & Video & Web & \xmark             & Human + AI & Annotated partial       & Video QA \\
\rowcolor[HTML]{EFEFEF}
\textbf{\benchmarkname (Ours)}         & 22K & Image & Web     & 183 locations & Human + AI & Standards-based        & Guidance + QA \\
\bottomrule
\end{tabular}}
\caption{Comparison of existing BLV guidance datasets. \benchmarkname uniquely offers geographic diversity across 183 locations, standards-based BLV involvement, and comprehensive task coverage at scale.}
\label{table:comparison_benchmark}
\end{table*}

Early BLV assistive approaches, including electronic travel aids and computer vision–based systems, primarily focus on obstacle detection and avoidance~\cite{kandalan2020techniques, lin2019deep}. While effective for basic safety, these methods struggle to capture richer contextual information required in complex real-world environments~\cite{liu2023open, xie2024emerging}. Recent advances in multimodal large language models (MLLMs)~\cite{liu2023visual, achiam2023gpt} offer new opportunities for higher-level scene understanding, and have begun to be explored for BLV assistance~\cite{xie2024emerging, merchant2024generating, zhao2024vialm, hwang2023system, kim2025understanding, han2025space}.

\begin{figure}[!t]
\includegraphics[width=0.48\textwidth]{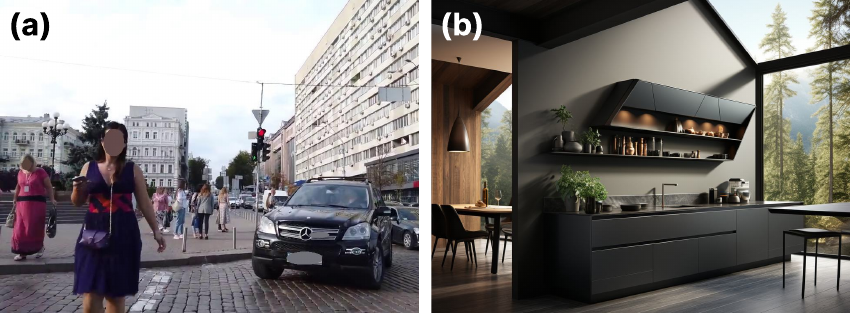}
\caption{Side-by-side comparison of examples from (a) \benchmarkname and (b) VIALM \cite{zhao2024vialm}. \benchmarkname consists of real-world scenes from a pedestrian viewpoint, while VIALM comprises home and supermarket images.}
\label{fig:sample_comparison}
\end{figure}

However, effective guidance data for BLV assistance remains difficult to obtain. Sighted annotators often struggle to anticipate BLV-specific needs, making them unsuitable for BLV-aware annotation~\cite{tigwell2021nuanced, islam2024identifying, morris2020ai}, and collecting diverse egocentric data requires costly manual exploration~\cite{han2025space}. Consequently, annotation relies on a small pool of domain experts~\cite{gurari2018vizwiz, zhao2024vialm}, resulting in BLV benchmarks that remain extremely small and limited in diversity, as summarized in \Cref{table:comparison_benchmark} and illustrated in \Cref{fig:sample_comparison}~\cite{zhao2024vialm, merchant2024generating}.

In this work, we introduce \benchmarkname, a large-scale accessibility-aware dataset for blind and low vision (BLV) guidance, grounded in structured standards distilled from established BLV guidelines (see \Cref{fig:standards_fig}). The dataset is constructed using a verification-centric human–AI pipeline that replaces free-form annotation with structured generation of \textit{silver} labels followed by human verification into high-fidelity \textit{gold} labels~\cite{kim2024meganno+,papadopoulos2016we}, enabling scalable data collection while maintaining annotation quality.

To capture realistic, in-the-wild scenarios reflective of BLV users' daily mobility experiences, \benchmarkname\ samples frames from geographically and visually diverse `walking videos'. These web-sourced videos are captured from an egocentric viewpoint across diverse real-world landscapes, mirroring the perspective of individuals with BLV. \benchmarkname contains a 22K-image dataset (including 2K human-verified data) covering a broad range of real-world settings, each annotated under \benchmarkname standards. Additionally, we construct \benchmarkQAname, an evaluation subset featuring multiple-choice question-answer pairs. As \benchmarkname contains real-world scenes, \benchmarkQAname enables fine-grained evaluation of visual perception, such as verifying which objects are truly present (from correct and incorrect options) and determining relative depths among detected objects, an essential capability for both BLV assistance systems and broader egocentric applications like robotics.

Using \benchmarkname, we evaluate several open-source and proprietary models. Our results show that proprietary models, such as GPT-4o, demonstrate superior zero-shot capabilities for BLV guidance. On \benchmarkQAname, open-source models lead in object recognition but lag significantly in depth perception compared to proprietary counterparts. Notably, fine-tuning with silver labels further boosts performance. These findings underscore that accurate spatial understanding is critical for effective BLV assistance. We hope these benchmarks foster further research into MLLMs for BLV assistive technologies and real-world visual perception.

Our contributions are:
\begin{enumerate}
  \item \benchmarkname, an accessibility-aware dataset with 22K real-world scene image–description pairs designed for blind and low vision (BLV) users, along with \benchmarkQAname, an 818-sample QA benchmark for evaluating fine-grained visual perception in real-world scenes.
  \item A scalable dataset construction pipeline that shifts annotation from generation to human verification, using \benchmarkname standards distilled from established BLV guidance.
  \item Experimental results demonstrating that spatial understanding and adherence to BLV-specific standards remain key challenges for current MLLMs.
\end{enumerate}

\section{\benchmarkname Dataset}
\label{sec:method}

We present \benchmarkname, an egocentric multimodal dataset designed to evaluate MLLMs on BLV guidance in diverse real-world environments. To ensure scalability without compromising quality, we employ a two-stage pipeline that prioritizes verification over generation (\Cref{subsec:dataset_construction}). First, an automated pipeline generates high-quality \textit{silver} labels guided by established \benchmarkname standards; then, trained annotators verify and refine these labels to produce authoritative \textit{gold} labels for evaluation following guidance from expert organizations and prior research.

\begin{figure*}[!t]
    \includegraphics[width=1\textwidth]{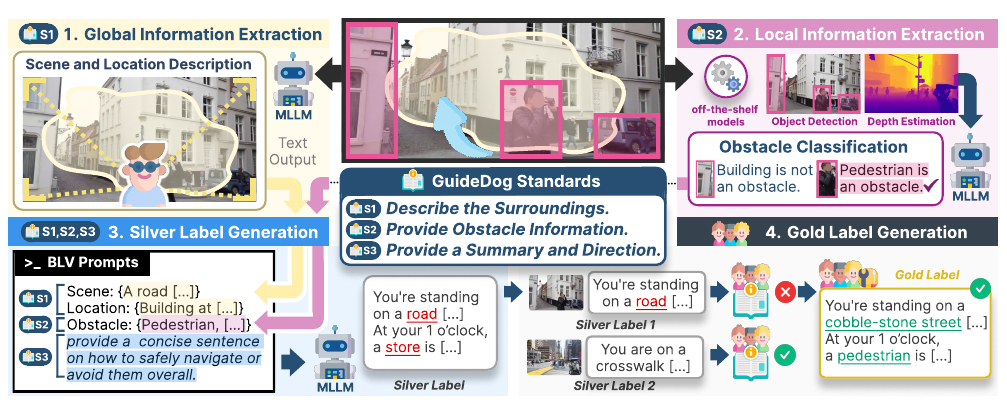}
    \caption{An overview of the \benchmarkname generation pipeline, ensuring all stages adhere to \benchmarkname standards (\SOne, \STwo, \SThree). For the collected scene image displayed at the center top, (1) in accordance with \SOne, the MLLM first extracts a comprehensive scene description; (2) following \STwo, both off-the-shelf models and the MLLM identify obstacles; (3) next, the extracted information is incorporated into a BLV-specific instruction designed to adhere to \SThree, generating a silver label that satisfies \SOne, \STwo, and \SThree; and (4) finally, human annotators assess and refine the silver labels to produce gold labels.}
    \label{fig:pipeline}
\end{figure*}

\subsection{\benchmarkname Standards for BLV Guidance}

\label{subsec:blv_standards}
Prior research has characterized BLV mobility challenges through 
extensive user studies~\cite{merchant2024generating, KURIAKOSE2023118720, 
song2010touch, duh2020v, manduchi2011mobility, madake2023qualitative}, 
while organizations have codified practical protocols for sighted 
guides~\cite{visionaustralia, senseguide, wsuaccess, wisconsin_dosdonts, 
wisconsin_guidetech, visionlossresources, stevens2003assisting, 
bemyeyes2024tips}. To structure these varied recommendations for 
MLLM evaluation, we consolidated them into three standards: \SOne, 
\STwo, and \SThree. 

\begin{tcolorbox}[
    colback=myLightBlue,
    colframe=myDarkDarkBlue,
    arc=2mm,
    left=2mm,
    right=2mm,
    top=3pt,
    bottom=2pt,
    boxsep=2pt,
    width=\linewidth,   %
]
    \begin{enumerate}[
        label=\textbf{S\arabic*}.,
        leftmargin=*,
        itemsep=2pt,  %
        parsep=0pt,
        topsep=3pt
    ]
        \item Describe the Surroundings.
        \item Provide Obstacle Information.
        \item Provide a Summary and Direction.
    \end{enumerate}
\end{tcolorbox}

\paragraph{\textit{S1. Describe the Surroundings.}} To orient BLV individuals in unfamiliar environments, assistive systems should provide contextual descriptions that establish spatial awareness~\cite{merchant2024generating, senseguide, wisconsin_guidetech, wisconsin_dosdonts}. This standard clearly identifies the user's current location and key environmental elements~\cite{KURIAKOSE2023118720, song2010touch}. For example, ``You are on a busy pedestrian street with shops on both sides and a clear path for walking surrounded by people.''

\paragraph{\textit{S2. Provide Obstacle Information.}} Obstacle awareness is a critical component of safe mobility for BLV individuals~\cite{islam2024identifying}. This standard focuses on delivering comprehensive information about obstacles, including their type, location, and proximity, enabling BLV individuals to make informed decisions about their path~\cite{KURIAKOSE2023118720, fernandes2019review, visionaustralia, senseguide, bemyeyes2024tips, wisconsin_dosdonts, wisconsin_guidetech}. For instance, ``Directly ahead at 12 o'clock, approximately 4 steps away, there is a sign post that could pose a risk of collision.''

\paragraph{\textit{S3. Provide a Summary and Direction.}} Cognitive load management is essential for BLV navigation. This standard promotes offering concise summaries with intuitive measurements (e.g., ``3 steps'', ``1 o'clock direction'') instead of precise measurements (e.g., ``3ft'',``5m'')~\cite{merchant2024generating, duh2020v, wisconsin_guidetech, bemyeyes2024tips}, while avoiding overly detailed explanations that may be overwhelming~\cite{merchant2024generating, bemyeyes2024tips}. For example, ``To navigate safely, proceed straight ahead with slight adjustments to avoid people at 12 and 2 o'clock.''

These standards serve as the foundation for our annotation pipeline, ensuring high-quality guidance faithful to BLV best practices while enabling scalable dataset creation. Per-standard references are provided in Appendix \ref{apdx:standards}.

\subsection{Dataset Construction}
\label{subsec:dataset_construction}

We present the data collection pipeline for \benchmarkname, which consists of four key stages: (a) Scene Image Collection, (b) Scene Information Extraction, (c) Silver Label Generation, and (d) Gold Label Generation. The latter three stages (b, c, d) are illustrated in ~\Cref{fig:pipeline}.

\paragraph{Scene Image Collection.}
Diverse real-world settings are crucial for representing the daily experiences of BLV individuals. To address this, we source egocentric `walking videos' from readily available YouTube channels that capture pedestrian perspectives. These channels typically provide extensive footage spanning multiple hours. 

Formally, let \(\mathcal{V}\) be the set of all videos. We apply a keyword-based filter to remove scenes irrelevant to BLV needs (e.g., those containing ``drive'', ``car'', or ``bike'') and eliminate non-relevant content such as shopping or vlogs, resulting in the filtered set of videos \(\mathcal{V}_{\text{filtered}}\).

We next discard videos shorter than six minutes. For the remaining videos, we remove the first five minutes and the final minute, which often contains intro and outro, and exclude any live or restricted videos. From the filtered video set, \(\mathcal{V}_{\text{filtered}}\), we then use GPT-4o \cite{achiam2023gpt} to extract city and region information for each video, allowing us to maintain a geographically diverse distribution. Specifically, we sample at most 5 videos per city and at most 10 videos per region, producing a final video set \(\mathcal{V}_{\text{sampled}}\).

From \(\mathcal{V}_{\text{sampled}}\), we first allocate a maximum of 50,000 total frames across videos based on their respective city-region distribution, ensuring balanced geographic coverage. For shorter videos, we sample at a minimum rate of one frame every 10 seconds to avoid overly redundant coverage. For longer videos, we adaptively increase the sampling interval so that the total sampled frames remain within the 50,000 limit. We then remove near-identical frames using DINO \cite{caron2021emerging} to reduce duplication further, resulting in a diverse collection of scene images for subsequent processing.

\begin{figure*}[!t]
    \includegraphics[width=1\textwidth]{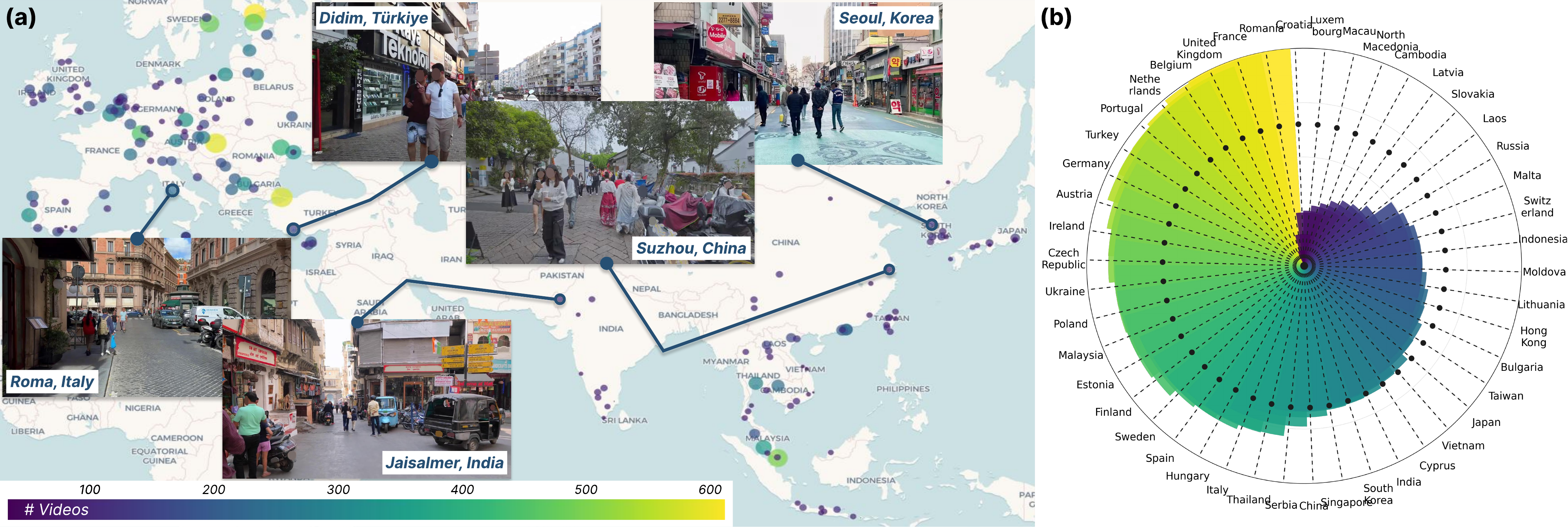}
    \caption{Visualization of the worldwide (a) city and (b) region distribution of samples in the \benchmarkname dataset.}
    \label{fig:data_distribution}
\end{figure*}

\paragraph{Scene Information Extraction.}
To generate \textit{silver} labels that comply with \benchmarkname standards, we extract two types of information: (1) global and (2) local. Each component provides information that enables MLLMs to generate \textit{silver} labels in accordance with \SOne and \STwo. %

For global information, we employ GPT-4o to filter irrelevant or noisy scenes (e.g., those where the camera points at the sky or ground or where the view is completely blocked). Then, GPT-4o extracts both a scene description \(\mathcal{T}^\text{s}\) and a location description \(\mathcal{T}^\text{l}\), capturing overall contextual details that satisfy \SOne. Formally, for the \(i\)-th scene, we obtain the global information \(\mathcal{X}_i^\text{global} = \bigl(t_{i}^s,\, t_{i}^l\bigr)\).

For local information, we extract key objects and their associated spatial properties (location, distance, direction) to enable MLLMs to generate \textit{silver} labels satisfying \STwo. MLLMs often struggle with fine-grained visual discrimination \cite{liang2024survey, jiao2024enhancing} and exhibit inconsistent object naming (synonyms). To address this limitation, we predefine a curated set of 80 crucial objects based on previous research \cite{islam2024identifying}, enabling more consistent and reliable detection of obstacles and hazards relevant to BLV navigation.

We employ an open-world object detector \cite{cheng2024yolo} to locate these objects in each \(i\)-th scene, yielding a set of \(k\) detected object classes \(\mathcal{O}_{i}\) and corresponding bounding boxes \(\mathcal{B}_{i}\). Next, a depth estimation model \cite{bochkovskii2024depth} generates a depth map \(m_i\) for the \(i\)-th scene. For each detected \(j\)-th object, its distance \(d_{ij}\) is computed as the median of depth values within the bounding box \(b_{ij}\). Note that these distances serve as auxiliary spatial signals; human verification relies on relative depth ordering between objects rather than absolute metric values.
To facilitate intuitive distance comprehension for BLV users, we convert these distances into step units (0.7\,m per step) \cite{sekiya1997optimal}. To determine each object's direction \(l_{ij}\), we assign a clock-face direction (10, 11, 12, 1, or 2 o'clock) to each bounding box center based on its horizontal position in the image.

From this process, we obtain the object set \(\mathcal{A}_i = \{(o_{ij}, b_{ij}, d_{ij}, l_{ij}) \mid j = 1,\dots,n\}\) for the \(i\)-th scene. Finally, we feed the scene \(i\) and its associated information \(\mathcal{A}_i\) into an MLLM to identify potentially crucial objects. Only those objects \(\mathcal{X}_i^\text{local}\) (a subset of \(\mathcal{A}_i\)) are used to generate the final \textit{silver} labels.

\paragraph{Silver Label Generation.}
In this stage, we combine the global information \(\mathcal{X}_i^\text{global}\) and filtered local objects \(\mathcal{X}_i^\text{local}\) with an accessibility-aware instruction for the MLLM to produce labels adhering to \SOne, \STwo, and \SThree. The resulting \textit{silver} labels thus conform to key \benchmarkname standards. We also apply EgoBlur~\cite{raina2023egoblur} to safeguard privacy by blurring individual faces and license plates in the original images.

\paragraph{Collecting Human Annotations.}

To obtain high-quality \textit{gold} labels for BLV guidance generation, we employ a human verification process for \textit{silver} labels.  Three sighted annotators verified and refined the \textit{silver} labels by filtering out unsuitable images and low-fidelity \benchmarkname standard annotations, while correcting any inaccuracies.

The verification applies two filters: image-level suitability ($c_1$) and standard-level adherence ($c_2$). Of the reviewed samples, $866$ ($26.5\%$) were rejected for image-level issues (e.g., unreadable frames, extreme angles, severe occlusion), and $265$ ($8.1\%$) for failing BLV guidance standards, confirming that the pipeline produces standards-compliant outputs when given suitable images.

Additionally, a randomly sampled subset of 150 images is annotated to construct \benchmarkQAname. The annotators validate the detected object classes and their calculated distances. From these verified objects, we create multiple-choice question-answer pairs for two distinct tasks: (1) object recognition and (2) relative depth comparison. For object recognition, each verified object was presented alongside three distractor objects randomly selected from the 80 predefined classes not present in the image. For relative depth comparison, we randomly selected pairs of verified objects from each image and created questions about which object was closer to or farther from the camera. The final benchmark consists of 435 questions across 150 images for object recognition, and 383 questions across 135 images for relative depth comparison.

\begin{table}[t!]
\centering
\resizebox{0.9\linewidth}{!}{
\begin{tabular}{lr}
\toprule
\# Source Videos & 269 \\
\# Total Frames in Source Videos & 59.8M \\
Total Source Videos Duration & 291 hours \\
\midrule
\# \benchmarkname Samples  & 22,084 \\
\# \benchmarkname Gold Label & 2,106  \\
\# \benchmarkQAname Samples & 818 \\
\midrule
\# Cities in \benchmarkname & 183 \\
\# Countries in \benchmarkname & 46 \\
\bottomrule
\end{tabular}}
\caption{\benchmarkname data statistics. “Source videos” indicates that images in \benchmarkname are sampled from video data.}
\label{table:dataset_analysis}
\end{table}

\subsection{Dataset Analysis}
\label{sec:dataset_analysis}
As presented in \Cref{table:dataset_analysis}, the \benchmarkname dataset comprises 22,084 samples in total, with 19,978 silver-labeled and 2,106 gold-labeled samples. \benchmarkQAname further provides 818 QA samples for evaluating fine-grained visual perception. These samples are curated from 269 source videos with a combined duration of 291 hours, representing approximately 59.8 million candidate frames. \benchmarkname can be easily scaled up by collecting more videos or sampling more frames.
The dataset is not only substantial in size but also diverse in geographic coverage, spanning 183 cities across 46 countries. Our pipeline implements a balanced region-city sampling strategy to prevent regional biases. As illustrated in ~\Cref{fig:data_distribution}, this approach prevents regional skews and produces a well-balanced global distribution. This geographic diversity ensures that \benchmarkname captures a wide range of environments, essential for developing robust BLV assistance systems that can function effectively across diverse real-world settings.

\begin{table*}[t!]
    \centering
    \resizebox{1.0\linewidth}{!}{
    \begin{tabular}{cl|cc|cc|cc|rr|cc|c}
    \toprule
    \multicolumn{1}{l}{\multirow{2}{*}{}} & \multirow{2}{*}{Model} & \multicolumn{2}{c|}{BLEU-2}     & \multicolumn{2}{c|}{BLEU-4}     & \multicolumn{2}{c|}{ROUGE-L}    & \multicolumn{2}{c|}{METEOR}     & \multicolumn{2}{c|}{GPT-Eval}   & \multicolumn{1}{c}{Gemini-Eval} \\
    \multicolumn{1}{l}{}                  &                        & 0-shot         & 3-shot         & 0-shot         & 3-shot         & 0-shot         & 3-shot         & 0-shot         & 3-shot         & 0-shot         & 3-shot         & 0-shot         \\
    \midrule
    \multirow{8}{*}{\rotatebox[origin=c]{90}{MLLM}} & Cambrian-1             & 0.218          & 0.329          & 0.087          & 0.175          & 0.221          & 0.323          & 0.267          & 0.375          & 0.219          & 0.307          & 0.277          \\
    & Molmo                  & 0.139          & 0.360          & 0.042          & 0.205          & 0.205          & 0.350          & 0.277          & 0.423          & 0.268          & 0.334          & 0.323          \\
    & LLaVA-1.6              & 0.128          & 0.320          & 0.048          & 0.173          & 0.226          & 0.313          & 0.311          & 0.352          & 0.149          & 0.276          & 0.322          \\
    & LLaVA-OneVision        & 0.183          & 0.343          & 0.058          & 0.180          & 0.225          & 0.337          & 0.300          & 0.404          & 0.264          & 0.334          & 0.350          \\
    & Qwen2.5-VL             & 0.179          & 0.347          & 0.069          & 0.190          & 0.228          & 0.341          & 0.294          & 0.412          & 0.230          & 0.319          & 0.330          \\
    & \quad \rotatebox[origin=c]{180}{$\Lsh$} w/ finetuning & \textbf{0.408} & {\underline {0.418}}    & \textbf{0.235} & {\underline {0.246}}    & \textbf{0.399} & \textbf{0.400} & \textbf{0.471} & 0.456          & \textbf{0.541} & \textbf{0.529} & 0.589          \\
    & Gemini 2.0 Flash       & 0.275          & 0.359          & 0.141          & 0.208          & 0.311          & 0.375          & 0.400          & {\underline {0.463}}    & 0.462          & 0.481          & \textbf{0.664} \\
    & GPT-4o                 & {\underline {0.339}}    & \textbf{0.425} & {\underline {0.188}}    & \textbf{0.252} & {\underline {0.359}}    & {\underline {0.399}}    & {\underline {0.445}}    & \textbf{0.474} & {\underline {0.490}}    & {\underline {0.505}}    & {\underline {0.651}} \\
    \midrule
    \multirow{2}{*}{\rotatebox[origin=c]{90}{SM}} & Gemini 2.0 Flash       & 0.324          & 0.324          & 0.172          & 0.173          & 0.341          & 0.342          & 0.435          & 0.435          & 0.281          & 0.289          & 0.588          \\
    & GPT-4o                 & 0.310          & 0.309          & 0.160          & 0.159          & 0.327          & 0.326          & 0.419          & 0.417          & 0.384          & 0.391          & 0.563          \\
    \bottomrule
    \end{tabular}}
    \caption{Experimental results of the guidance generation task on \benchmarkname. The best scores are \textbf{boldfaced}, and the second-best scores are \underline{underlined}.}
    \label{table:main_results}
\end{table*}

\section{Task Definition}
\label{sec:task}
We propose two complementary tasks to evaluate models' capabilities in assisting BLV individuals: (1) an accessibility-aware guidance generation task that provides comprehensive navigational assistance and (2) a real-world scene understanding task that assesses fundamental visual perception abilities through multiple-choice questions. The first task evaluates holistic BLV guidance according to established standards, while the second targets specific perceptual skills essential for effective navigation. We evaluate these tasks using our \benchmarkname and \benchmarkQAname datasets, respectively.

\paragraph{Guidance Generation.}
This task requires models to generate comprehensive guidance for BLV users from a given image, following the \benchmarkname standards. To elicit standardized responses, we ask models to: (1) summarize the surroundings (\SOne), (2) concisely describe hazards using clock-face directions and step counts (\STwo), and (3) provide a final direction (\SThree).

\begin{figure}[!t]
    \includegraphics[width=0.48\textwidth]{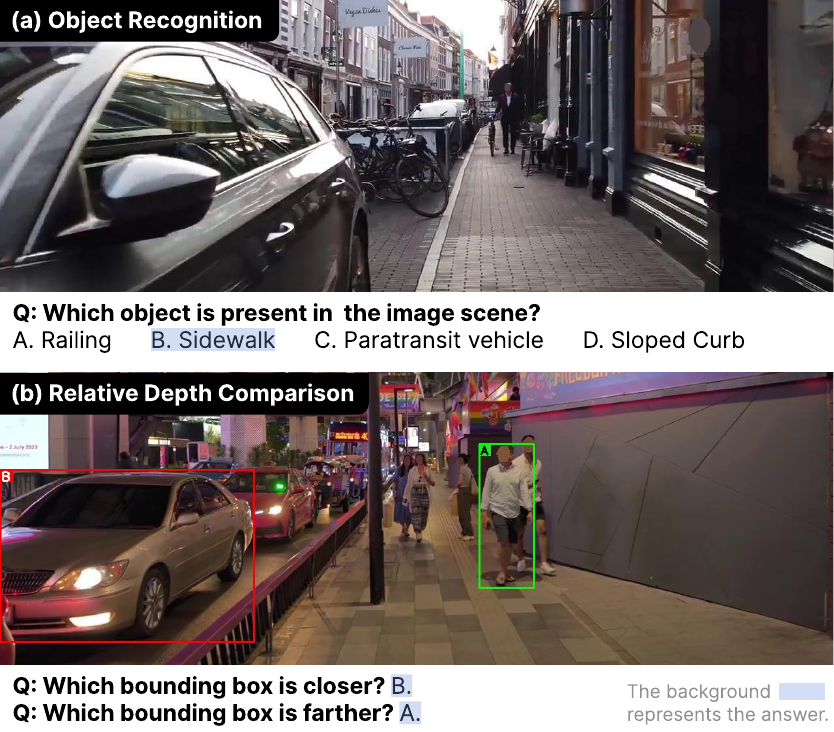}
    \caption{Examples of object recognition and relative depth comparison questions from \benchmarkQAname.}
    \label{fig:GuideDogQA_sample}
\end{figure}

\paragraph{Visual Perception QA.}
To diagnose failure modes in guidance generation and benchmark real-world scene perception, we assess two perceptual skills via multiple-choice questions. For object recognition, models select which of four candidate objects actually appears in the image. For depth comparison, models determine which of two bounding-boxed objects is closer or farther. Examples are shown in~\Cref{fig:GuideDogQA_sample}.

\section{Experiments}
\label{sec:result}

We evaluate models on two tasks: accessibility-aware guidance generation on \benchmarkname and visual perception QA on \benchmarkQAname. We first describe the evaluated models and then present results for each task.

\paragraph{Models}
We evaluate a diverse set of vision-language systems, including open-source, proprietary, and Socratic models~\cite{zengsocratic}.
For \textit{open-source} models, we consider LLaVA 1.6~\cite{liu2024improved}, LLaVA-OneVision~\cite{li2024llava}, Qwen-2.5-VL~\cite{bai2025qwen2}, Cambrian-1~\cite{tong2025cambrian}, and Molmo~\cite{deitke2024molmo}. LLaVA 1.6, LLaVA-OneVision, and Qwen-2.5-VL represent high-performing open-source approaches, Cambrian-1 features a distinctive multi-encoder architecture, and Molmo is notable for its transparency. Additionally, we fine-tune Qwen-2.5-VL using LoRA \cite{hu2022lora} on our silver labels to investigate the impact of fine-tuning on performance.
Among \textit{proprietary models}, we include GPT-4o~\cite{achiam2023gpt} and Gemini 2.0 Flash~\cite{google2024gemini2}, both recognized for their exceptional performance on general vision-language tasks.

Finally, we implement a Socratic Model (SM)~\cite{zengsocratic} using a two-stage framework: first, converting visual inputs into intermediate textual descriptions, and then performing text-only reasoning based on those descriptions and the associated object information \(\mathcal{A}_i\) generated by off-the-shelf models (as described in~\Cref{subsec:dataset_construction}). For image caption generation, we use LLaVA-1.6, while reasoning is performed using Gemini 2.0 Flash or GPT-4o. This approach enables us to assess the role of direct visual perception.

\subsection{Accessibility-aware Guidance Generation}
We evaluate models on the guidance generation task using \benchmarkname, focusing on how well the generated outputs align with the \benchmarkname standards.

\paragraph{Metric}
Detailed performance metrics across all models and settings are shown in \Cref{table:main_results}.
To assess the alignment between model-generated outputs and ground-truth text, we employ standard natural language generation metrics, including BLEU-2, BLEU-4~\cite{papineni2002bleu}, ROUGE-L~\cite{lin2004rouge}, and METEOR~\cite{banerjee2005meteor}.

To capture high-level semantic alignment with the \benchmarkname standards beyond surface-level text similarity, we use LLM-based evaluation~\cite{liu2023gpteval} with GPT-4o as an automated evaluator to assess guidance quality based on BLV-specific criteria. Additionally, we conduct reference-based evaluation using GPT-4o to compare generated guidance against gold labels. GPT-4o measures the degree to which generated outputs align with the corresponding gold labels, and we report accuracy based on its assessments. To mitigate evaluator bias, we further validate with Gemini 2.0 Flash as an alternative evaluator, yielding consistent model rankings.

\begin{table}[t!]
    \centering
    \resizebox{0.92\linewidth}{!}{
    \begin{tabular}{lcc}
    \toprule
    Model & \makecell{Depth \\ Comparison} & \makecell{Object \\ Recognition} \\
    \midrule
    Random Chance & 25.0 & 25.0 \\
    \midrule
    Cambrian-1             & 24.3          & 82.3          \\
    Molmo                  & 28.5          & 34.0          \\
    LLaVA-1.6              & 30.0          & 80.9          \\
    LLaVA-OneVision        & 32.4          & \textbf{87.4} \\
    Qwen2.5-VL             & 22.2          & {\underline {85.7}}    \\
    \quad \rotatebox[origin=c]{180}{$\Lsh$} w/ finetuning & 41.5 & 83.9 \\
    Gemini 2.0 Flash       & {\underline {53.0}}    & 65.7          \\
    GPT-4o                 & \textbf{67.1} & 74.7          \\ 
    \bottomrule
    \end{tabular}}
    \caption{Experimental results of the object recognition and relative depth comparison tasks on \benchmarkQAname.}
    \label{table:task2}
\end{table}

\begin{table}[t!]
    \centering
    \resizebox{0.85\linewidth}{!}{
    \begin{tabular}{lcccc}
    \toprule
    Model & S1 & S2 & S3 & Avg \\
    \midrule
    Cambrian-1 &            3.06    &   1.92    &   2.15 & 2.38 \\
    Molmo &                 3.02    &   2.41    &   2.36 & 2.60 \\
    LLaVA-OneVision &       3.26    &   2.47    &   2.49 & 2.74 \\
    Qwen2.5-VL &            2.93    &   2.27    &   2.23 & 2.48 \\
    \quad \rotatebox[origin=c]{180}{$\Lsh$} w/ finetuning & 3.64 & {\underline {3.45}} & {\underline {3.67}} & 3.59 \\
    Gemini 2.0 Flash & {\underline {3.95}} & {\underline {3.45}} & 3.40 & {\underline {3.60}} \\
    GPT-4o & \textbf{4.15} & \textbf{3.8} & \textbf{3.76} & \textbf{3.90} \\
    \midrule
    (SM) GPT-4o &       2.88    &   2.02    &   2.7  & 2.53 \\
    \midrule
    Filtered Silver & 4.76 & 4.62 & 4.52 & 4.63 \\
    \bottomrule
    \end{tabular}}
    \caption{Results of the user study evaluated using a Likert scale ranging from 1 to 5, where 1 indicates poor adherence to the \benchmarkname standard and 5 indicates high adherence. Best is \textbf{boldfaced}; second-best is \underline{underlined}.}
    \label{table:human_study}
\end{table}

\paragraph{Visual Perception Matters for Guidance.}
GPT-4o exhibits the strongest zero-shot capabilities, while open-source models struggle significantly, with LLaVA-1.6 yielding the lowest scores across most metrics.
SMs achieve competitive performance despite the absence of direct visual input processing, but remain inferior to MLLMs. This suggests that both direct visual perception and a robust language model foundation are essential for generating effective BLV guidance.

\paragraph{In-context Learning Helps to Adapt to Guidance Generation.} In-context learning through few-shot examples substantially improves performance across nearly all models.  In contrast, SM approaches show minimal performance changes between zero-shot and 3-shot settings. This suggests that in-context learning helps models understand the guidance generation task, while SM models already possess this understanding.

\paragraph{Fine-tuning Is Beneficial.} Fine-tuning on BLV-specific data yields dramatic performance improvements, outperforming all other models. %
Moreover, fine-tuned Qwen2.5-VL demonstrates remarkable stability across zero-shot and few-shot settings, with only minimal additional gains from in-context examples. This suggests that fine-tuning on domain-specific data effectively embeds the BLV guidance principles that would otherwise need to be communicated through examples.

\subsection{Question Answering}
We evaluate models on the visual perception QA task using \benchmarkQAname, which measures object recognition and relative depth comparison.

\paragraph{Metric}
We report accuracy on the multiple-choice QA task, which evaluates models' ability to identify objects present in the image and compare their relative depth. For depth comparison, we consider the answer correct only if the model correctly identifies both the closer and the farther objects to reduce the language prior. Detailed performance metrics for both object recognition and depth comparison tasks are presented in \Cref{table:task2}.

\paragraph{Open-source Models Recognize Objects Well.} For object recognition, open-source models demonstrate surprisingly strong performance. Interestingly, proprietary models like GPT-4o and Gemini 2.0 Flash underperform in this task despite their superior guidance generation capabilities. The fine-tuned Qwen2.5-VL maintains strong object recognition performance, showing only a slight decrease ($-1.8\%$) from its base model while substantially improving depth comparison ($+19.3\%$, see \Cref{table:task2}). This asymmetry suggests that occasional object-level noise (primarily hallucinated or misdetected objects in silver labels) is the main quality bottleneck of the silver-label pipeline, rather than structural or standard-adherence issues.
\paragraph{Depth Comparison Remains a Challenge.} Depth perception presents a significant challenge across all models, especially for open-source models. Cambrian-1 and Qwen2.5-VL could not surpass random chance. The fine-tuned Qwen2.5-VL shows notable improvement in depth perception, indicating that BLV-specific training enhances spatial reasoning capabilities.
This finding aligns with the needs of BLV users, who require not just information about what objects are present, but also their spatial relationships and proximity. 

\subsection{User Study}
To complement our automated metrics, we conduct a subjective user study to assess how well different models adhere to the \benchmarkname standards. As shown in \Cref{table:human_study}, we recruited 14 participants to evaluate model-generated guidance across all three \benchmarkname standards (\SOne: Describe the Surroundings, \STwo: Provide Obstacle Information, and \SThree: Provide Summary and Direction).
The participants rated the outputs of each model on a 5-point Likert scale \cite{likert1932technique}, where a score of 1 indicates poor adherence to the standard and a score of 5 indicates excellent adherence. We randomly sampled 86 images from the gold labels and collected guidance outputs from each model under evaluation. 

\paragraph{Results.} Our human evaluation results align with the automated metrics while providing additional insights into model performance across specific standards. GPT-4o consistently received the highest ratings across all three standards. Gemini 2.0 Flash ranks second overall, comparable to fine-tuned Qwen2.5-VL. For reference, silver labels evaluated by the same protocol achieved $4.76$ (\SOne), $4.62$ (\STwo), and $4.52$ (\SThree).
The relatively low scores for \STwo across most models highlight the challenge of accurately describing obstacles with appropriate spatial references that align with our depth perception results in \Cref{table:task2}.
Socratic GPT-4o performs worse than GPT-4o, confirming our earlier finding that visual information is crucial for effective BLV guidance. Additionally, the Socratic approach performed relatively better on \SThree, suggesting that summary and guidance may rely more on high-level reasoning than on fine-grained visual details.

\section{Related Works}
\label{sec:related_work}

\paragraph{MLLMs for BLV Assistance.}
Recent multimodal large language models (MLLMs)~\cite{achiam2023gpt, liu2023visual, google2024gemini2, bai2025qwen2} have advanced vision-language capabilities beyond object recognition toward holistic and contextual scene understanding. Building on these advances, several works have explored MLLM-based assistance for blind and low vision (BLV) individuals, including navigation guidance~\cite{zhao2024vialm, merchant2024generating, xie2024emerging}, object localization~\cite{liu2024objectfinder}, robotic guide dog systems~\cite{hwang2023system, kim2025understanding, han2025space}, and broader evaluations of MLLMs as visual assistants for BLV users~\cite{karamolegkou-etal-2025-evaluating}. However, most existing approaches rely on small-scale datasets collected from limited domains, constraining their generalization to real-world BLV mobility scenarios.

\paragraph{Egocentric and BLV Datasets.}
Large-scale egocentric datasets such as Ego4D~\cite{grauman2022ego4d} advance first-person scene understanding but are not tailored to BLV needs. BLV-specific efforts include ORBIT~\cite{massiceti2021orbit}, an object recognition dataset directly captured by BLV users, and ViewQA~\cite{song2024video}, which offers VQA with a 360-degree egocentric camera. EgoBlind~\cite{xiao2025egoblind} further collects BLV-recorded egocentric videos from the web for visual assistance. VizWiz~\cite{gurari2018vizwiz, gurari2019vizwiz, huh2024long} pioneered accessibility-aware VQA, while VIALM~\cite{zhao2024vialm} and \citet{merchant2024generating} explored guidance generation, though both remain limited in scale due to costly expert annotation. \citet{an2025can} studied BLV user preferences for navigation. WalkVLM~\cite{yuan2024walkvlm} contributes a video-based BLV walking dataset validated through surveys with visually impaired participants. \benchmarkname instead grounds its standards in BLV guidelines and prior studies, and scales annotation through a verification-centric human-AI pipeline to $22$K samples from $183$ cities across $46$ countries.

\section{Conclusion}
\label{sec:conclusion}

We introduce \benchmarkname, a scalable accessibility-aware benchmark for evaluating MLLMs on BLV guidance against established mobility standards. Our analysis reveals that current models struggle to adhere to BLV-specific standards and lack the spatial understanding critical for safe navigation. We hope this work catalyzes future research into better spatial and temporal understanding, real-time navigation systems, and personalized assistance to better serve the global BLV community.

\section{Limitations}
\label{sec:limitation}

\paragraph{Static Images}
We use static images to isolate guidance generation quality. Video inputs introduce confounds such as query timing and temporal coherence. Some hazards, including escalators and revolving doors, are motion-defined and may appear safe in a static frame; capturing such implicit motion blindness~\cite{zhang2025escalator} requires temporally grounded evaluation. Because \benchmarkname is sourced from continuous walking tours, it can be extended to video-based navigation in future work.

\paragraph{Cross-Cultural Spatial Language}
Although \benchmarkname spans 46 countries and reflects regional differences in object distributions and infrastructure (\Cref{sec:regional_analysis}), our annotations rely on universal spatial references and do not model cross-cultural variation in spatial language or object semantics~\cite{karamolegkou-etal-2024-vision}. Modeling such variation remains an important direction for future work.

\section{Ethical Considerations}

\paragraph{License \& Privacy}
We restricted data collection to videos published under the Creative Commons (CC-BY-SA) license. To safeguard privacy in public spaces, we applied EgoBlur~\cite{raina2023egoblur} to automatically detect and blur faces and license plates, followed by a human verification stage to ensure complete anonymization.

\paragraph{Societal Impact}
\benchmarkname aims to advance assistive technologies for the BLV community. We recognize that current MLLMs exhibit hallucinations and explicitly advise against deploying models trained on this data as standalone navigational aids without additional safety guardrails and extensive real-world validation.

\section*{Acknowledgments}
This work was supported by the AI Seoul Tech Research Support Program of the Seoul Future Foundation. This work was partly supported by Institute of Information \& Communications Technology Planning \& Evaluation (IITP) grant funded by the Korean Government (MSIT) (No.~RS-2021-II211343, Artificial Intelligence Graduate School Program (Seoul National University)), the National Research Foundation of Korea (NRF) grant funded by the Korea government (MSIT) (No.~RS-2024-00354218), and the Technology Innovation Program (RS-2025-25456760, Development of a humanoid robot specialized in chemical processes based on AI foundation model) funded by the Ministry of Trade, Industry and Resources (MOTIR, Korea). We thank the KAIT GPU project for computational support. The ICT at Seoul National University provides research facilities for this study.

\bibliography{custom}

\clearpage
\appendix
\section{References Supporting the GuideDog Standards for BLV Guidance}
\label{apdx:standards}

This section lists the key references that inform the development of the \benchmarkname Standards for BLV Guidance in \Cref{subsec:blv_standards}. The supporting literature for each standard (S1–S3) is organized below.

\paragraph{S1. Describe the Surroundings}
\cite{merchant2024generating, senseguide, wisconsin_guidetech, wisconsin_dosdonts, KURIAKOSE2023118720, song2010touch}

\paragraph{S2. Provide Obstacle Information}
\cite{merchant2024generating, KURIAKOSE2023118720, fernandes2019review, visionaustralia, senseguide, wsuaccess, wisconsin_dosdonts, wisconsin_guidetech}

\paragraph{S3. Provide a Summary and Direction}
\cite{KURIAKOSE2023118720, fernandes2019review, visionaustralia, senseguide, wsuaccess, wisconsin_dosdonts, wisconsin_guidetech, merchant2024generating, bemyeyes2024tips}

\section{Automatic Pipeline Details}
\subsection{Gathering Videos}
To construct a high-quality dataset of outdoor walking scenes, we implemented a systematic approach to video selection. We manually identified YouTube channels specializing in walking tours across urban and natural environments. All selected channels (listed in \Cref{table:table_apdx_youtube_channels}) satisfy three criteria: (1) content exclusively featuring walking tours, (2) availability under Creative Commons (CC) licensing, and (3) diverse geographical coverage.
Following channel identification, we developed an automated pipeline to extract relevant video URLs. This pipeline is initialized by parsing channel metadata from a JSON configuration file and establishing a structured directory hierarchy for organizing filtered results.
Our filtering mechanism applies multiple criteria to ensure content quality and relevance: (1) Exclusion of titles containing non-walking activities through regex-based keyword filtering (e.g., ``drive'', ``car'', ``drone'', ``shopping'', ``market'', ``VR''); (2) Removal of videos shorter than six minutes to ensure sufficient content depth; (3) Automatic skipping of live broadcasts and restricted content. After preprocessing, the resulting data distributions are visualized in \Cref{fig:data_distribution} using OpenStreetMap\footnote{\url{https://www.openstreetmap.org/}} Carto style\footnote{\url{https://carto.com/platform}} and Plotly\footnote{\url{https://plotly.com/python/maps/}}.

\begin{figure*}[!t]
    \centering
    \includegraphics[width=1.0\linewidth]{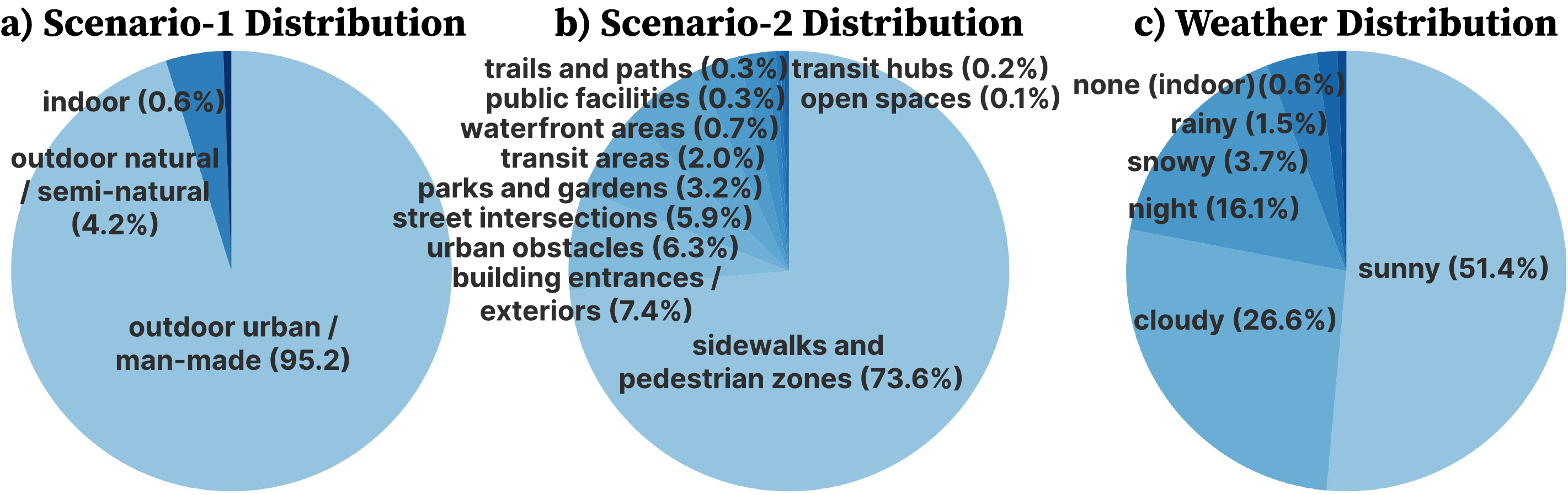}
    \caption{Scene distribution in the \benchmarkname dataset based on a Places365-inspired taxonomy (Level-1 and Level-2 categories), along with overall weather category distribution.}
    \label{fig:x_data_distribution}
\end{figure*}

\subsection{Scene Information Extraction}
To generate silver labels, we extract both global and local scene information using GPT-4o. For global extraction, the model first validates whether an input image represents a suitable street scene by providing a binary "Yes" or "No" assessment. Images are disqualified if (1) large objects obstruct the majority of the frame, (2) the focal point emphasizes elements peripheral to the street environment (e.g., store displays, signage), or (3) the camera perspective is oriented significantly upward or downward rather than at pedestrian level.  
After validation, the model produces a comprehensive scene description encompassing pedestrians, buildings, vehicles, and other salient elements. The spatial arrangement is documented using a clock-position reference system limited to the 10, 11, 12, 1, and 2 o'clock positions (with 10 o'clock corresponding to the leftmost portion of the image, 12 o'clock to the center, and 2 o'clock to the rightmost area). The prompt is shown in \Cref{fig:prompt_apdx_global}.

For local information extraction, we employ an off-the-shelf object detection model combined with distance estimation. Detected objects are categorized into two zones based on a 5-meter proximity threshold. In the Complete Danger Zone, we evaluate whether objects directly intersect the user's projected walking path, potentially resulting in a collision. In the Ordinary Zone, we focus primarily on dynamic elements such as approaching vehicles (motorcycles, cars), bicycles, and pedestrians. To refine this classification, we prompt the MLLM to provide explicit reasoning for each object's danger designation. The prompt is shown in \Cref{fig:prompt_apdx_local}.

\subsection{Silver Label Generation}
To generate silver labels that adhere to \benchmarkname standards, we integrate the extracted global context information and filtered local object data using a specialized prompt designed to satisfy all three standards (\SOne, \STwo, and \SThree), as shown in \Cref{fig:prompt_apdx_silver}. During this stage, we also implement EgoBlur~\cite{raina2023egoblur} technology to ensure privacy protection by automatically blurring faces and license plates in the original images.

\subsection{Collecting Human Annotations}
\begin{figure*}[!t]
    \centering
    \includegraphics[width=0.75\textwidth]{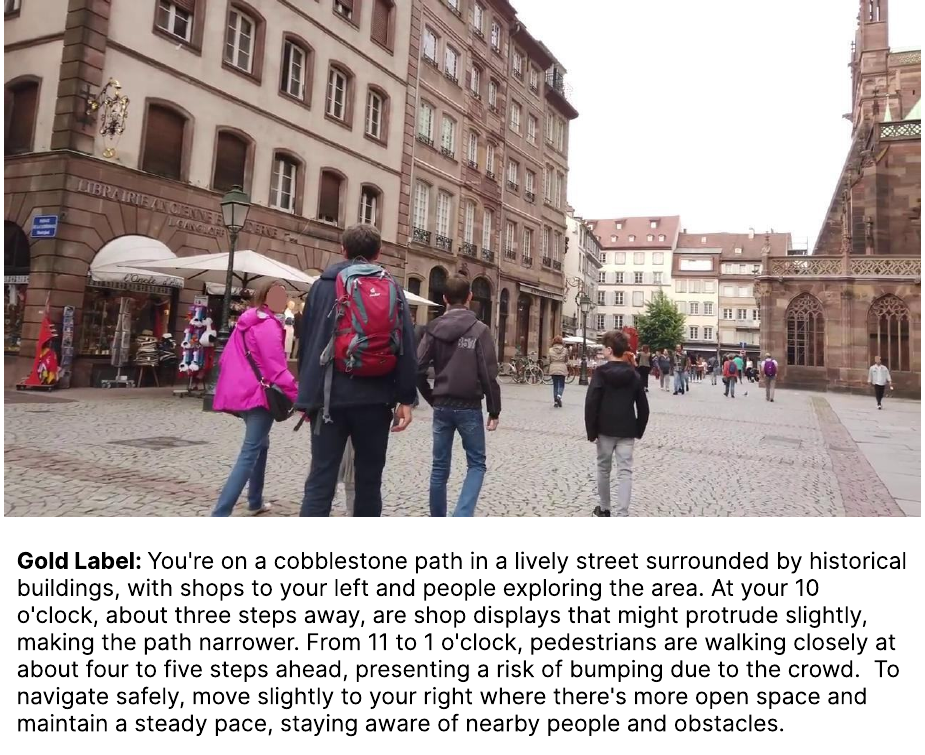}
    \caption{An example of a gold label in \benchmarkname.}
    \label{fig:datapoint_apdx_gold}
\end{figure*}

\begin{figure*}[!t]
    \centering
    \includegraphics[width=0.75\textwidth]{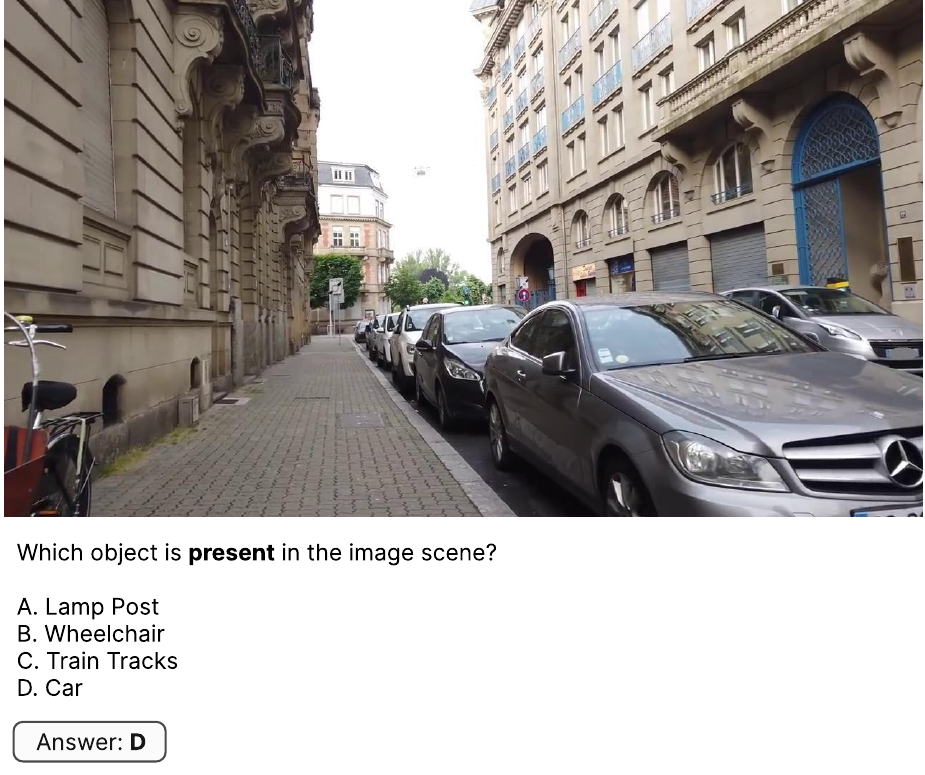}
    \caption{An example of an object recognition task on \benchmarkQAname.}
    \label{fig:datapoint_apdx_object}
\end{figure*}

\begin{figure*}[!t]
    \centering
    \includegraphics[width=0.75\textwidth]{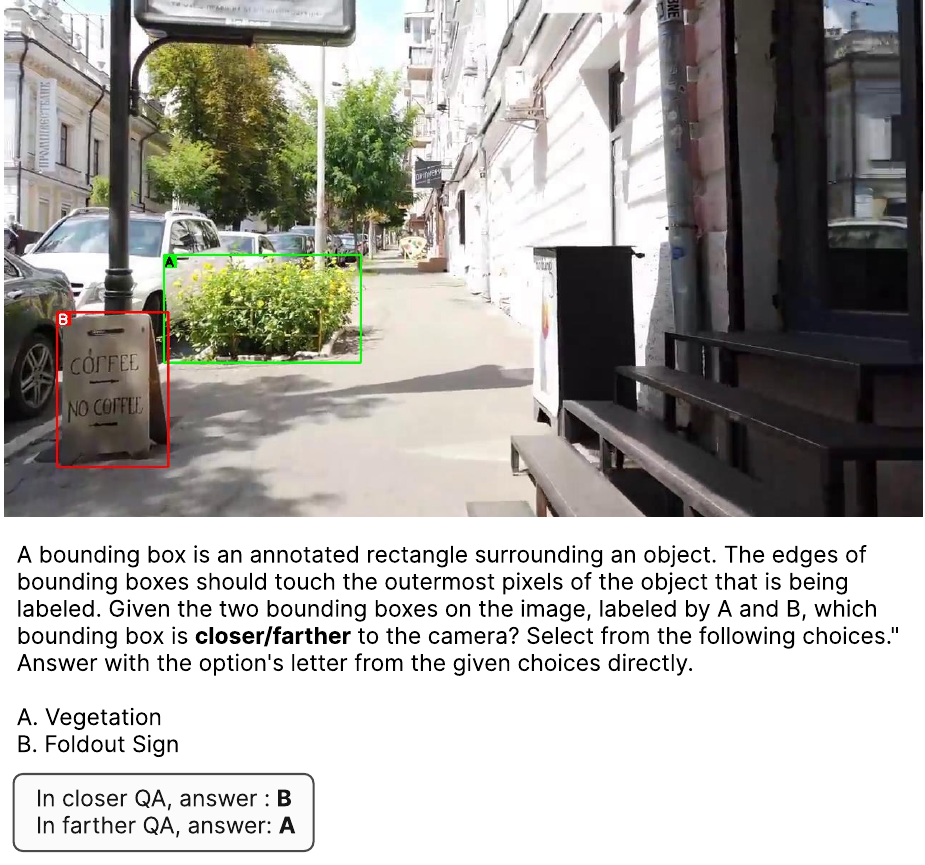}
    \caption{An example of a depth recognition task on \benchmarkQAname.}
    \label{fig:datapoint_apdx_depth}
\end{figure*}

To establish gold-standard labels, we implemented a rigorous human verification protocol. Three sighted annotators reviewed the silver labels using Label Studio tools \cite{LabelStudio} (UI shown in \Cref{fig:generate_goldlabel}), filtering out unsuitable images and substandard annotations while correcting any inaccuracies. To ensure fair compensation, we paid the dedicated annotators approximately 1.5 times the local minimum hourly wage. Separately, the subjective user study was conducted with student volunteers.
For \benchmarkQAname construction, we randomly sampled 150 images from the verified dataset. Annotators validated both object classifications and distance measurements using the interface depicted in \Cref{fig:depth_validation}. From these validated objects, we developed two categories of multiple-choice questions: (1) object recognition tasks, where each verified object is presented alongside three randomly selected distractors from the 80 predefined classes not present in the image; and (2) relative depth comparison tasks, where randomly selected object pairs prompt questions about relative proximity to the camera. The final benchmark comprises 435 object recognition questions across 150 images and 383 depth comparison questions spanning 135 images. Example gold labels for both \benchmarkname and \benchmarkQAname are illustrated in Figures \ref{fig:datapoint_apdx_gold}, \ref{fig:datapoint_apdx_object}, and \ref{fig:datapoint_apdx_depth}.

\begin{table*}[t!]
\centering
\resizebox{0.8\linewidth}{!}{
\begin{tabular}{ll}
\toprule

YouTube Channels          & Channel URLs                                              \\ \midrule
Roma Walking Tour                  & \url{https://www.youtube.com/@RomaWalkingTour}     \\
DuckTravel (World Tourist Channel) & \url{https://www.youtube.com/@DuckTravel}       \\
Realistic Tourist                  & \url{https://www.youtube.com/@RealisticTourist} \\
POPtravel                          & \url{https://www.youtube.com/@poptravelorg}

\\ \bottomrule
\end{tabular}}
\caption{The list of YouTube channels from which videos were downloaded.}
\label{table:table_apdx_youtube_channels}
\end{table*}

\section{Dataset Details}
\subsection{Fine-grained Analyses}

We expand Section 2.3 by providing fine-grained dataset statistics. These include scene complexity (average of 9.3 objects per image) and description analysis (average length of 126 words, range: 52–212).  As shown in \Cref{fig:x_data_distribution}, we analyzed scene and scenario distributions using a Places365-inspired taxonomy. These were categorized and validated by GPT-4.1. One author manually verified 30 samples, confirming high agreement between GPT and human annotations (e.g., Level-1 scenario: 96\% vs. 98\%; Level-2 scenario: 86\% vs. 86\%; weather: 92\% vs. 94\%).

Regarding object distribution, beyond the most frequent category, person (29.5\%), we observe a significant presence of vehicles (14.8\%, including car, bus, bicycle) and other traffic- or street-related objects such as foldout signs (7.2\%), lamp posts (5.5\%), barrier posts (3.6\%), and benches (2.0\%). This distribution varies across regions and scenes, for example, vehicles are more common at street intersections (28.1\%). Notably, in Vietnam, motorcycles (24.8\%) significantly outnumber cars (13.2\%).

\subsection{Regional Object Distribution}
\label{sec:regional_analysis}
To investigate whether \benchmarkname reflects regional variation rather than averaging it out, we analyze object distributions across two regional groupings: East/Southeast Asia and four European subregions.

\begin{table}[h!]
\centering
\small
\resizebox{\linewidth}{!}{
\begin{tabular}{lcc}
\toprule
Object & \makecell{Southeast Asia \\ ($n=186$)} & \makecell{East Asia \\ ($n=106$)} \\
\midrule
Motorcycle    & 42\% & 8\%  \\
Car           & 59\% & 25\% \\
Foldout Sign  & 40\% & 58\% \\
Road          & 33\% & 20\% \\
Street Vendor &  6\% &  3\% \\
\bottomrule
\end{tabular}}
\caption{Object frequency in Southeast Asia vs.\ East Asia samples.}
\label{table:apdx_regional_asia}
\end{table}

Southeast Asia exhibits a vehicle-dominant environment, with motorcycles dominant in Vietnam (77\%), Cambodia (74\%), and Thailand (50\%). In contrast, East Asia shows higher prevalence of foldout signs (58\% vs.\ 40\%), indicating signage-dense pedestrian environments.

\begin{table}[h!]
\centering
\small
\resizebox{\linewidth}{!}{
\begin{tabular}{lcccc}
\toprule
Object & \makecell{W.\ Europe \\ ($n=611$)} & \makecell{N.\ Europe \\ ($n=279$)} & \makecell{E.\ Europe \\ ($n=503$)} & \makecell{S.\ Europe \\ ($n=253$)} \\
\midrule
Foldout Sign     & 45\% & 28\% & 34\% & 30\% \\
Bicycle          & 32\% & 11\% &  9\% &  8\% \\
Trash bins       & 33\% & 16\% & 22\% & 21\% \\
Building         & 57\% & 72\% & 51\% & 57\% \\
Lamp Post        & 34\% & 41\% & 45\% & 37\% \\
Sidewalk         & 33\% & 25\% & 29\% & 38\% \\
Raised Entryway  & 15\% & 13\% &  9\% & 16\% \\
\bottomrule
\end{tabular}}
\caption{Object frequency across European subregions.}
\label{table:apdx_regional_europe}
\end{table}

Western Europe shows a strong bicycle culture (32\%; Netherlands 48\%), while Southern Europe exhibits higher sidewalk (38\%) and raised entryway (16\%) frequencies, suggesting elevation-related walking risks. These differences indicate that \benchmarkname structurally reflects meaningful regional variations in mobility culture and urban infrastructure.

\subsection{Viewpoint Comparison with BLV-Recorded Video}
\label{sec:viewpoint_analysis}
A natural concern with using sighted walking-tour footage is the gap between this viewpoint and that of a BLV user with a head-mounted assistive camera. To quantify this gap, we computed optical-flow-based motion statistics on \benchmarkname source videos and on EgoBlind~\cite{xiao2025egoblind}, a BLV-recorded video dataset. We sampled one matched 20-second window per video ($N=332$ per domain) and extracted frames at $2$~FPS for dense optical-flow analysis.

\begin{table*}[h!]
\centering
\begin{tabular}{lcccc}
\toprule
Metric & KS stat & $p$-value & Cliff's $d$ & Effect size \\
\midrule
Flow Magnitude (median)         & 0.273 & $<\!0.001$ & $-0.329$ & Small \\
Global Motion (mean)            & 0.253 & $<\!0.001$ & $-0.281$ & Small \\
Global Motion (std / jitter)    & 0.134 & $\phantom{<}0.005$ & $\phantom{-}0.118$ & Negligible \\
Stationary Ratio                & 0.278 & $<\!0.001$ & $\phantom{-}0.333$ & Medium \\
Bottom-1/3 Motion (mean)        & 0.391 & $<\!0.001$ & $-0.480$ & Large \\
\bottomrule
\end{tabular}
\caption{Optical-flow motion statistics comparing \benchmarkname source videos and EgoBlind ($N=332$ per domain).}
\label{table:apdx_optical_flow}
\end{table*}

EgoBlind exhibits a higher stationary ratio (median $16.6\%$ vs.\ $8.3\%$), indicating more frequent static or slow-movement content. \benchmarkname shows substantially stronger bottom-region motion (Cliff's $d = -0.48$, large), consistent with sustained forward walking and ground-plane dynamics. Jitter differences are negligible, suggesting comparable camera stability during active movement.

To explain these quantitative differences, we annotated 50 random clips per domain:

\begin{table}[h!]
\centering
\small
\resizebox{\linewidth}{!}{
\begin{tabular}{lcc}
\toprule
Property & \makecell{EgoBlind \\ ($n=50$)} & \makecell{\benchmarkname \\ ($n=50$)} \\
\midrule
Standing            & 21 (42\%) &  0 (0\%)  \\
Slow walk           & 26 (52\%) & 42 (84\%) \\
Normal walk         &  1 (2\%)  &  7 (14\%) \\
Cane visible        & 21 (42\%) &  0 (0\%)  \\
Edit/cut visible    & 15 (30\%) &  0 (0\%)  \\
Subtitle occlusion  & 30 (60\%) &  2 (4\%)  \\
\bottomrule
\end{tabular}}
\caption{Qualitative properties of 50 randomly sampled clips per domain.}
\label{table:apdx_qualitative_clips}
\end{table}

EgoBlind's higher stationary ratio reflects frequent standing ($42\%$) and slow walking ($52\%$), with visible cane motion absent in \benchmarkname. \benchmarkname predominantly shows slow walking ($84\%$), typical of walking tour recordings. EgoBlind clips also contain video edits ($30\%$) and subtitle overlays ($60\%$), both largely absent in \benchmarkname. Both datasets use body-mounted cameras ($98\%$), indicating that the motion differences stem from navigation behavior rather than camera configuration.

These results confirm a measurable viewpoint gap, which we acknowledge as a limitation. The two datasets are complementary: EgoBlind captures authentic BLV motion patterns but is constrained by edits, limited scale, and limited geographic diversity, while \benchmarkname offers clean, geo-diverse scenes aligned with smart-glasses-style assistive scenarios.

\section{Experiment Details}
\subsection{Training details}
To perform a more comprehensive analysis of \benchmarkname, we fine-tune Qwen-2.5-VL using LoRA on our silver labels. Specifically, we utilize Hugging Face AutoTrain\footnote{\url{https://huggingface.co/docs/autotrain/en/index}} with a learning rate of 2e-5, a batch size of 2, and train for 1 epoch, taking approximately 1 hour on four NVIDIA A6000 GPUs.

\subsection{Evaluation Details}
In this study, we conduct experiments using the \texttt{lmms-eval} repository\footnote{\url{https://github.com/EvolvingLMMs-Lab/lmms-eval}}. All evaluations are performed on a machine equipped with NVIDIA A100 GPUs in a single run.

\subsection{Evaluation Prompts}
We evaluate two main model types: the general MLLM and the SM model. For each model, we configured both 0-shot and 3-shot prompts. \Cref{fig:prompt_apdx_guidedog}, \Cref{fig:prompt_apdx_guidedog_few}, \Cref{fig:prompt_apdx_guidedog_sm}, and \Cref{fig:prompt_apdx_guidedog_sm_few} show these four configurations, respectively.
Additionally, we adopted GPT-Eval to semantically evaluate the outputs of these models. We designed the prompt based on \citet{yu2023mm} to assign scores ranging from 0 to 1, as shown in \Cref{fig:prompt_apdx_guidedog_gpteval}.

\subsection{User Study}
\begin{figure*}[!t]
    \centering
    \includegraphics[width=0.75\textwidth]{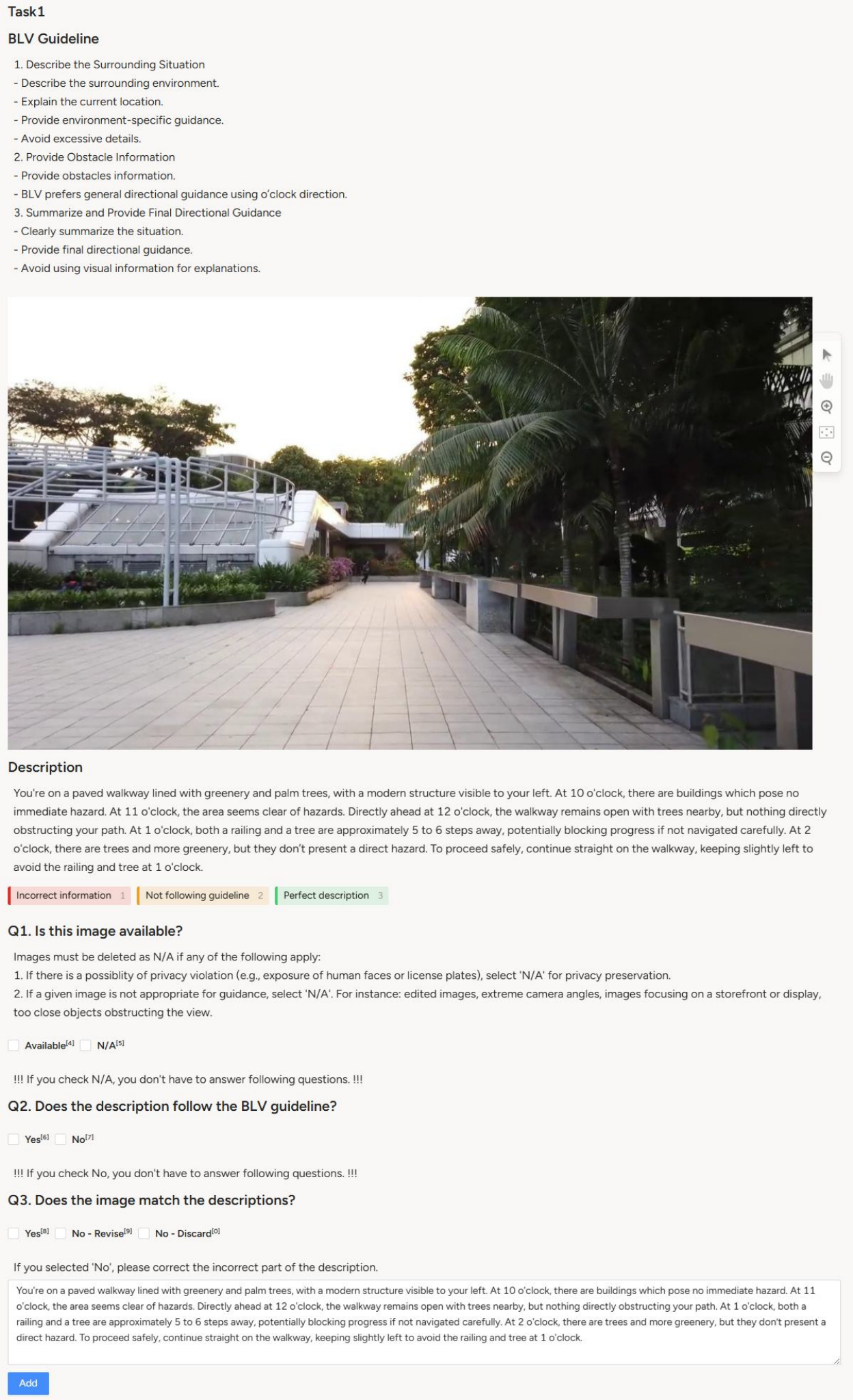}
    \caption{An interface for evaluating and refining silver labels into gold labels.}
    \label{fig:generate_goldlabel}
\end{figure*}

\begin{figure*}[!t]
    \centering
    \includegraphics[width=0.85\textwidth]{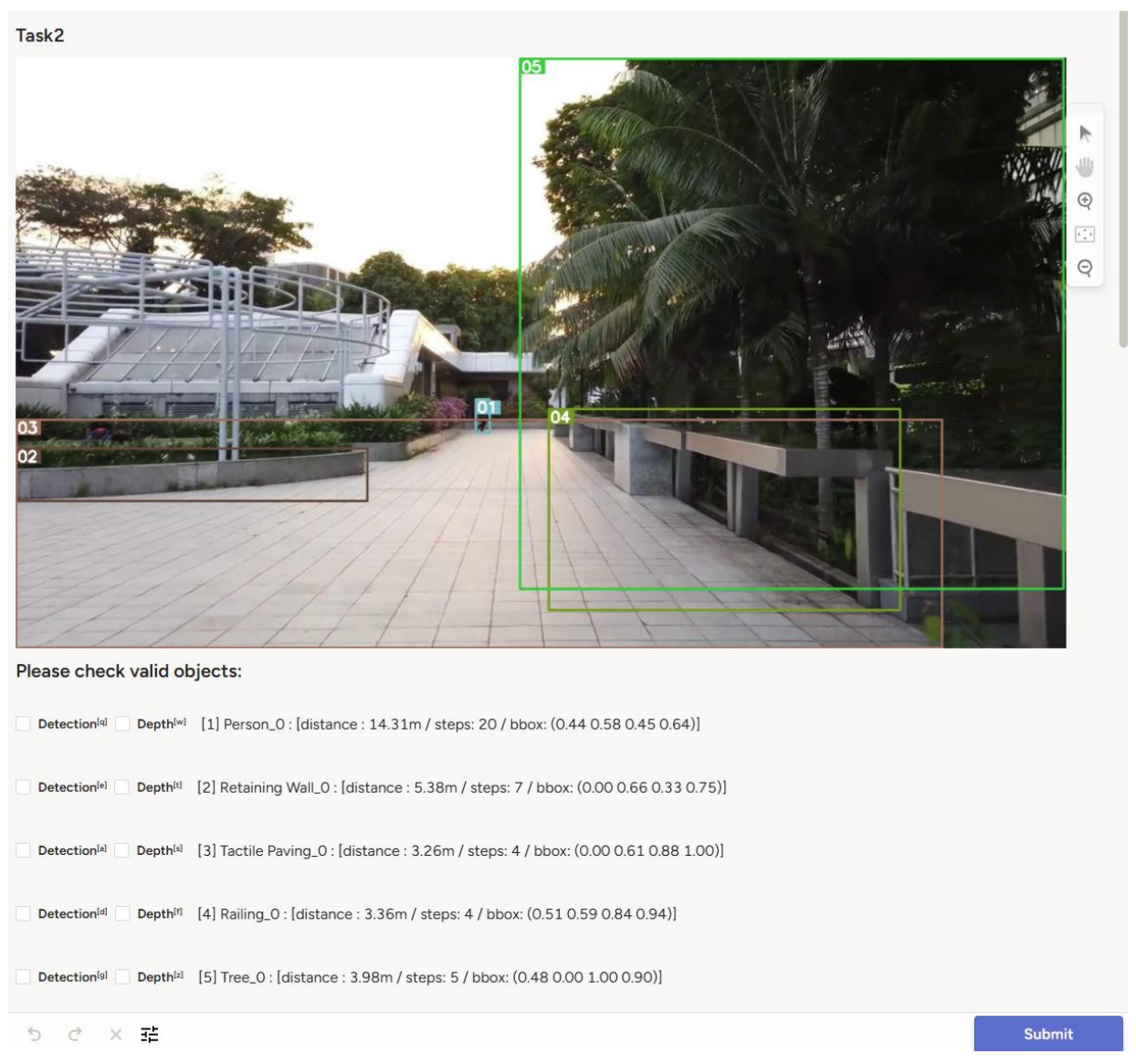}
    \caption{An interface for validating object detection and depth estimation.}
    \label{fig:depth_validation}
\end{figure*}

\begin{figure*}[!t]
    \centering
    \includegraphics[width=0.85\textwidth]{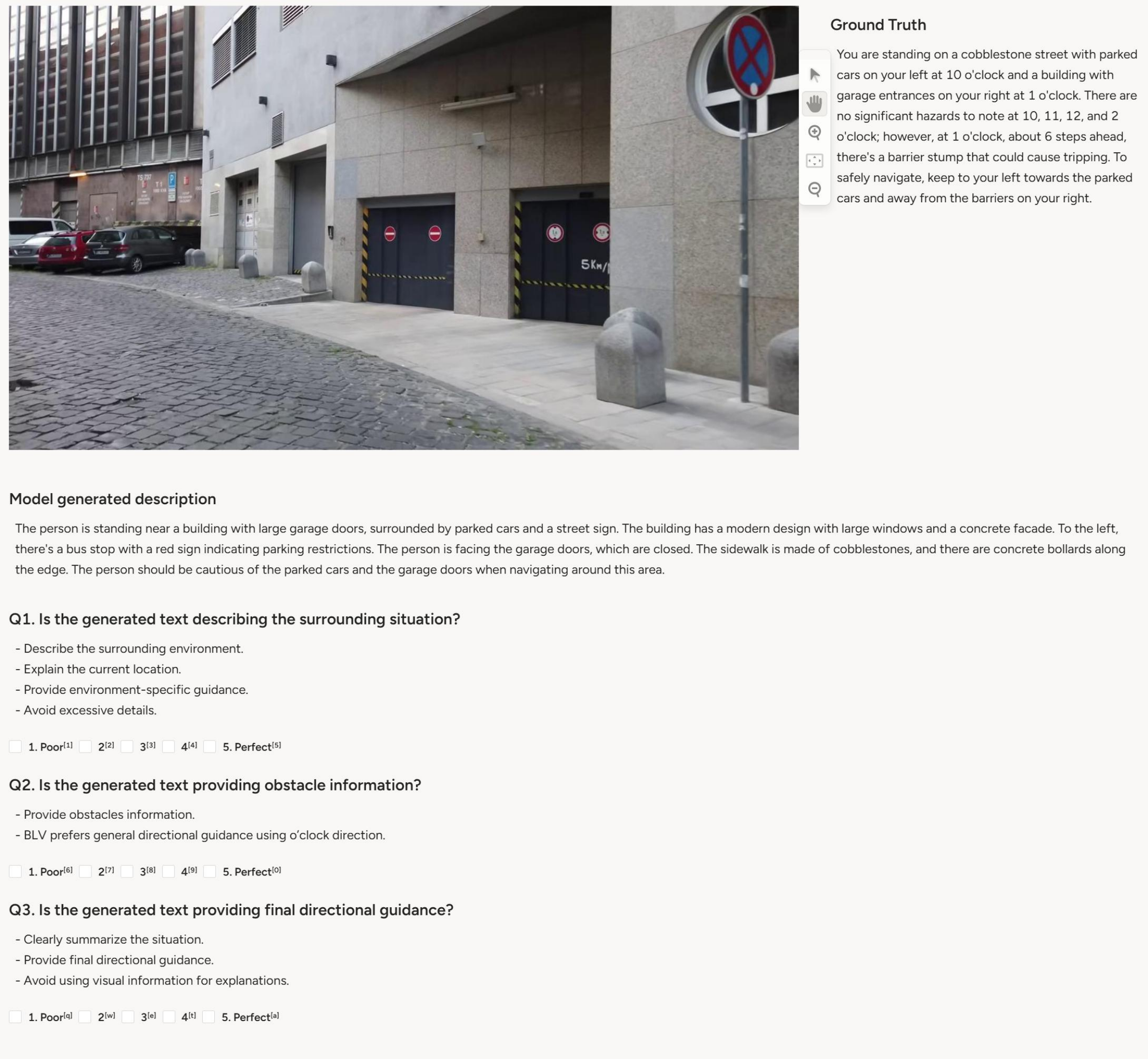}
    \caption{An interface for human evaluation of model-generated descriptions.}
    \label{fig:human_study}
\end{figure*}

We conduct a subjective user study to assess how well different models adhere to the \benchmarkname standards. We employ 14 human annotators to score the outputs of seven models per image using the Label Studio tool, as shown in \Cref{fig:human_study}. We construct a gold label set for \benchmarkname and \benchmarkQAname, and sample annotations are provided in \Cref{fig:generate_goldlabel} and \Cref{fig:depth_validation}.

\begin{figure*}
\centering
\small
\begin{tcolorbox}[
    colback=white, %
    colframe=gray, %
    arc=4mm %
]
You are an expert tasked with determining whether an image depicts a \textbf{street scene}.
Please follow the instructions below:\\
1. Review the image and decide if it represents a street scene ("Yes") or not ("No").

Filter out the following non-street cases from the label \texttt{\{is\_street\}}:
\begin{itemize}
    \item Certain objects block most of the screen, making it difficult to recognize the street.
    \item The viewer is looking at something other than the street (e.g., looking at the display glass of a store on the street, looking at a store's sign, etc.).
    \item The viewer is looking upwards or at the floor, not a straight sight of pedestrians.
\end{itemize}

And include the following cases:
\begin{itemize}
    \item If the image is of a pedestrian's sight, even if it does not depict a designated street or a clear path, because this viewer is blind or has low vision, they could unknowingly enter a roadway.
\end{itemize}

2. If \texttt{"Yes"}:
\begin{itemize}
    \item scene\_description: Provide an overview of the street, including pedestrians, buildings, vehicles, or any key elements that make it a street.
    \item scene\_location: Describe the location and surroundings using only 10, 11, 12, 1, and 2 o’clock positions. The leftmost part of the image is 10 o’clock, the center is 12 o’clock, and the rightmost is 2 o’clock.
\end{itemize}

3. If \texttt{"No"}:
\begin{itemize}
    \item scene\_description: Briefly explain why this is not a street scene.
    \item scene\_location: Must be \texttt{"None"}.
\end{itemize}

4. Use only the JSON format below.
5. Do not include any text outside of this JSON format.\

\vspace{0.3cm}\
\noindent Output JSON example:\
\begin{verbatim}
{
    "is_street": "Yes",
    "scene_description": "A detailed description of the street, including key
    elements such as pedestrians, shops, and vehicles.",
    "scene_location": "A positional overview using 10, 11, 12, 1, and 2 o’clock
    references."
}

Or:

{
    "is_street": "No",
    "scene_description": "Reason why this is not a street scene (e.g., highway, 
    indoor, or empty road).",
    "scene_location": "None"
}
\end{verbatim}

You must include exactly these three fields:\
\begin{itemize}
    \item \texttt{"is\_street"}
    \item \texttt{"scene\_description"}
    \item \texttt{"scene\_location"}
\end{itemize}\

No additional text beyond this JSON structure is allowed.\
\end{tcolorbox}
\caption{The global information extraction prompt filters out inappropriate images and extracts scene information and location details in accordance with \SOne. The inputs to the prompts are \textbf{boldfaced}. }
\label{fig:prompt_apdx_global}
\end{figure*}

\begin{figure*}
\centering
\small
\begin{tcolorbox}[
    colback=white, %
    colframe=gray, %
    arc=4mm %
]
You are an AI vision safety expert specializing in assisting visually impaired and low-vision users. Your task is to analyze a street image and determine which objects pose a danger to the user.\\

The object information is divided into two sections based on a distance threshold of 5 meters:\\

Complete Danger Zone (within 5 meters):\\
\textbf{\{in\_object\_info\}}\\

Ordinary Zone (beyond 5 meters):\\
\textbf{\{out\_object\_info\}}\\

\noindent Instructions:\
\begin{enumerate}
    \item Complete Danger Zone (within 5 meters):
    \begin{itemize}
        \item Evaluate whether each object is directly in the user's walking path and could lead to a collision.
        \item Mark an object as dangerous ("Yes") only if it poses a collision risk (e.g., curbs, potholes, poles, stairs, construction barriers, parked vehicles, etc.).
        \item Otherwise, mark it as not dangerous ("No") and briefly explain why.\\
    \end{itemize}
    
    \item Ordinary Zone (beyond 5 meters)
    \begin{itemize}
        \item Focus on moving objects in this zone (e.g., approaching motorcycles, cars, bicycles, pedestrians).
        \item Mark an object as dangerous ("Yes") if it is moving toward the user and could pose a threat.
        \item Otherwise, mark it as not dangerous ("No") and provide a brief explanation.
    \end{itemize}
\end{enumerate}

For each object, provide a JSON entry with the following three fields:\
\begin{itemize}
    \item \texttt{"object"}: the object's identifier or name.
    \item \texttt{"is\_dangerous"}: "Yes" if dangerous, "No" if not.
    \item \texttt{"why\_dangerous"}: a clear explanation for your assessment.
\end{itemize}

\noindent Output your entire response strictly in JSON format following this example:\
\begin{verbatim}
[
  {
    "object": "object_name",
    "is_dangerous": "Yes",
    "why_dangerous": "This object is in the user's path and may cause a 
    collision."
  },
  {
    "object": "object_name",
    "is_dangerous": "No",
    "why_dangerous": "This object is stationary and not in the user's immediate 
    path."
  }
]
\end{verbatim}

\noindent Do not include any text outside of the JSON structure.\
\end{tcolorbox}
\caption{The local information extraction prompt, which evaluates detected objects and classifies each as either dangerous or not, in order to satisfy \STwo. The inputs to the prompts are \textbf{boldfaced}.
}
\label{fig:prompt_apdx_local}
\end{figure*}

\begin{figure*}
\centering
\begin{tcolorbox}[
    colback=white, %
    colframe=gray, %
    arc=4mm %
]
You are an expert guide for visually impaired individuals. Your task is to provide a concise explanation based on the following guidelines, delivering the content as if speaking naturally without section breaks.\\

Guidelines:\
1) Surroundings and Position: Summarize where the person is, the general environment, their current position, and any nearby landmarks in 1-2 sentences.\\
2) Hazards:\
   \begin{itemize}
       \item For each direction (10, 11, 12, 1, and 2 o’clock), combine all hazards in that direction into exactly one sentence, mentioning approximate distance(s) and reason(s) they are dangerous.
       \item Follow the order of 10, 11, 12, 1, and 2 o’clock.
   \end{itemize}
3) Navigation: After describing all hazards, provide a single, concise sentence on how to safely navigate or avoid them overall.

Potential Hazards:\\
\textbf{\{object\_info\}}\\

Scene Information:\\
\textbf{\{scene\_info\}}\\

Remember to provide a single, flowing explanation without labeled sections, as if talking directly to the visually impaired individual.\\

\end{tcolorbox}
\caption{The silver label generation prompt leverages classified dangerous objects and extracted scene information to generate silver labels. The input of the prompts are \textbf{boldfaced}.}
\label{fig:prompt_apdx_silver}
\end{figure*}

\begin{figure*}
\begin{tcolorbox}[
    colback=white, %
    colframe=gray, %
    arc=4mm %
]
Compare the ground truth and prediction from AI models, to give a correctness score for the prediction. \textless AND\textgreater\ in the ground truth means it is totally right only when all elements in the ground truth are present in the prediction, and \textless OR\textgreater\ means it is totally right when any one element in the ground truth is present in the prediction. The correctness score is 0.0 (totally wrong), 0.1, 0.2, 0.3, 0.4, 0.5, 0.6, 0.7, 0.8, 0.9, or 1.0 (totally right). Just complete the last space of the correctness score.\\

gpt\_query\_prompt \textbar Ground truth \textbar Prediction \textbar Correctness\\
What is x in the equation? \textbar -1 \textless AND\textgreater -5 \textbar x = 3 \textbar 0.0 \\
What is x in the equation? \textbar -1 \textless AND\textgreater -5 \textbar x = -1 \textbar\ 0.5 \\
What is x in the equation? \textbar -1 \textless AND\textgreater -5 \textbar x = -5 \textbar 0.5 \\
What is x in the equation? \textbar -1 \textless AND\textgreater -5 \textbar x = -5 or 5 \textbar 0.5 \\
What is x in the equation? \textbar -1 \textless AND\textgreater -5 \textbar x = -1 or x = -5 \textbar 1.0 \\

Can you explain this meme? \textbar This meme is poking fun at the fact that the names of the countries Iceland and Greenland are misleading. Despite its name, Iceland is known for its beautiful green landscapes, while Greenland is mostly covered in ice and snow. The meme is saying that the person has trust issues because the names of these countries do not accurately represent their landscapes. \textbar The meme talks about Iceland and Greenland. It’s pointing out that despite their names, Iceland is not very icy and Greenland isn’t very green. \textbar 0.4 \\

Can you explain this meme? \textbar This meme is poking fun at the fact that the names of the countries Iceland and Greenland are misleading. Despite its name, Iceland is known for its beautiful green landscapes, while Greenland is mostly covered in ice and snow. The meme is saying that the person has trust issues because the names of these countries do not accurately represent their landscapes. \textbar The meme is using humor to point out the misleading nature of Iceland’s and Greenland’s names. Iceland, despite its name, has lush green landscapes while Greenland is mostly covered in ice and snow. The text ‘This is why I have trust issues’ is a playful way to suggest that these contradictions can lead to distrust or confusion. The humor in this meme is derived from the unexpected contrast between the names of the countries and their actual physical characteristics. \textbar 1.0 \\

Write a single, natural-sounding passage with no section breaks. Begin by providing a 1–2 sentence summary of the person's surroundings and position, including any nearby landmarks. Then, for each direction from 10 to 2 o'clock, combine all hazards in exactly one sentence, specifying approximate distances and why they are dangerous in the order of 10, 11, 12, 1, and 2 o'clock. Finally, offer a single concise sentence explaining how to safely navigate or avoid all hazards. 
\textbar \textbf{\{gold\_label\}} \textbar \textbf{\{model\_generated\_output\}}

\end{tcolorbox}
\caption{The GPT-Eval prompt, which evaluates how well it adheres to \benchmarkname standards. The inputs for the prompt are \textbf{boldfaced}.}
\label{fig:prompt_apdx_guidedog_gpteval}
\end{figure*}

\begin{figure*}
\centering
\begin{tcolorbox}[
    colback=white, %
    colframe=gray, %
    arc=4mm %
]
You are an expert guide for visually impaired individuals. Your task is to provide a concise explanation based on the following guidelines, delivering the content as if speaking naturally without section breaks.\\

Guidelines:\
1) Surroundings and Position: Summarize where the person is, the general environment, their current position, and any nearby landmarks in 1-2 sentences.\\
2) Hazards:\
   \begin{itemize}
       \item For each direction (10, 11, 12, 1, and 2 o’clock), combine all hazards in that direction into exactly one sentence, mentioning approximate distance(s) and reason(s) they are dangerous.
       \item Follow the order of 10, 11, 12, 1, and 2 o’clock.
   \end{itemize}
3) Navigation: After describing all hazards, provide a single, concise sentence on how to safely navigate or avoid them overall.\\

Remember to provide a single, flowing explanation without labeled sections, as if talking directly to the visually impaired individual.\\

\end{tcolorbox}
\caption{The zero-shot prompt designed for MLLM to generate accessibility-aware guidance generation in \benchmarkname. The inputs of the prompts are \textbf{boldfaced}.}
\label{fig:prompt_apdx_guidedog}
\end{figure*}

\begin{figure*}
\centering
\begin{tcolorbox}[
    colback=white, %
    colframe=gray, %
    arc=4mm %
]
You are an expert guide for visually impaired individuals. Your task is to provide a concise explanation based on the following guidelines, delivering the content as if speaking naturally without section breaks.\\

Guidelines:\
1) Surroundings and Position: Summarize where the person is, the general environment, their current position, and any nearby landmarks in 1-2 sentences.\\
2) Hazards:\
   \begin{itemize}
       \item For each direction (10, 11, 12, 1, and 2 o’clock), combine all hazards in that direction into exactly one sentence, mentioning approximate distance(s) and reason(s) they are dangerous.
       \item Follow the order of 10, 11, 12, 1, and 2 o’clock.
   \end{itemize}
3) Navigation: After describing all hazards, provide a single, concise sentence on how to safely navigate or avoid them overall.\\

Examples:
\begin{itemize}
    \item You're on a bustling city street with buildings on your left, and the sidewalk and storefronts on your right. At 10 o'clock, about five steps away, there's a moving car which is potentially dangerous if you stray off the sidewalk. To navigate safely, stay on the sidewalk, maintaining a safe distance from the road.
    \item You're standing in a lively marketplace with stalls under umbrellas at 12 o'clock and buildings in the background; there are parked vehicles to your sides. At 10 o'clock, approximately 5 steps away, there's a parked car that could obstruct any movement in that direction. At 11 o'clock, there are no immediate hazards. Directly ahead, at 12 o'clock, the market stalls might pose a minor obstacle if you walk too closely. At 1 o'clock, there is a car about 4 steps away, posing a potential obstacle. At 2 o'clock, no significant hazards are present. To navigate safely, proceed slowly towards 12 o'clock while veering slightly to your right to avoid the car at 1 o'clock.
    \item You are in a public plaza with buildings directly ahead at 12 o'clock and an art structure at 2 o'clock, with pedestrians around. At 10 o'clock, there is a barrier post and a pole about 3 steps away which could obstruct your path. At 11 o'clock, trash bins are 4 steps away, which might be a tripping hazard. Directly ahead at 12 o'clock, a foldout sign is 3 steps away, posing a risk of collision. At 2 o'clock, a barrier post is 4 steps away, which could also cause a trip. To safely navigate the area, move slightly to your left and proceed forward, avoiding the central obstacles.
\end{itemize}

Remember to provide a single, flowing explanation without labeled sections, as if talking directly to the visually impaired individual.\\

\end{tcolorbox}
\caption{The 3-shot prompt designed for MLLM to generate accessibility-aware guidance generation in \benchmarkname. The inputs of the prompts are \textbf{boldfaced}.}
\label{fig:prompt_apdx_guidedog_few}
\end{figure*}

\begin{figure*}
\centering
\begin{tcolorbox}[
    colback=white, %
    colframe=gray, %
    arc=4mm %
]
You are an expert guide for visually impaired individuals. Your task is to provide a concise explanation based on the following guidelines, delivering the content as if speaking naturally without section breaks.\\

Guidelines:\
1) Surroundings and Position: Summarize where the person is, the general environment, their current position, and any nearby landmarks in 1-2 sentences.\\
2) Hazards:\
   \begin{itemize}
       \item For each direction (10, 11, 12, 1, and 2 o’clock), combine all hazards in that direction into exactly one sentence, mentioning approximate distance(s) and reason(s) they are dangerous.
       \item Follow the order of 10, 11, 12, 1, and 2 o’clock.
   \end{itemize}
3) Navigation: After describing all hazards, provide a single, concise sentence on how to safely navigate or avoid them overall.\\

Remember to provide a single, flowing explanation without labeled sections, as if talking directly to the visually impaired individual.\\

Scene Description: \textbf{\{llava\_output}\}\\
Object Info:\\
\textbf{\{object\_info}\} \\
Guidance: 

\end{tcolorbox}
\caption{The zero-shot prompt designed for SM to generate accessibility-aware guidance generation in \benchmarkname. The inputs of the prompts are \textbf{boldfaced}.}
\label{fig:prompt_apdx_guidedog_sm}
\end{figure*}

\begin{figure*}
\centering
\small
\begin{tcolorbox}[
    colback=white, %
    colframe=gray, %
    arc=4mm %
]
You are an expert guide for visually impaired individuals. Your task is to provide a concise explanation based on the following guidelines, delivering the content as if speaking naturally without section breaks.\\

Guidelines:\\
1) Surroundings and Position: Summarize where the person is, the general environment, their current position, and any nearby landmarks in 1-2 sentences.\\
2) Hazards:\
   \begin{itemize}
       \item For each direction (10, 11, 12, 1, and 2 o’clock), combine all hazards in that direction into exactly one sentence, mentioning approximate distance(s) and reason(s) they are dangerous.
       \item Follow the order of 10, 11, 12, 1, and 2 o’clock.
   \end{itemize}
3) Navigation: After describing all hazards, provide a single, concise sentence on how to safely navigate or avoid them overall.\\

Remember to provide a single, flowing explanation without labeled sections, as if talking directly to the visually impaired individual.\\

Scene Description:\\
Object Info:\\
Guidance: You're on a bustling city street with buildings on your left, and the sidewalk and storefronts on your right. At 10 o'clock, about five steps away, there's a moving car which is potentially dangerous if you stray off the sidewalk. To navigate safely, stay on the sidewalk, maintaining a safe distance from the road.\\

Scene Description:\\
Object Info:\\
Guidance: You're standing in a lively marketplace with stalls under umbrellas at 12 o'clock and buildings in the background; there are parked vehicles to your sides. At 10 o'clock, approximately 5 steps away, there's a parked car that could obstruct any movement in that direction. At 11 o'clock, there are no immediate hazards. Directly ahead, at 12 o'clock, the market stalls might pose a minor obstacle if you walk too closely. At 1 o'clock, there is a car about 4 steps away, posing a potential obstacle. At 2 o'clock, no significant hazards are present. To navigate safely, proceed slowly towards 12 o'clock while veering slightly to your right to avoid the car at 1 o'clock.\\

Scene Description:\\
Object Info:\\
Guidance: You are in a public plaza with buildings directly ahead at 12 o'clock and an art structure at 2 o'clock, with pedestrians around. At 10 o'clock, there is a barrier post and a pole about 3 steps away which could obstruct your path. At 11 o'clock, trash bins are 4 steps away, which might be a tripping hazard. Directly ahead at 12 o'clock, a foldout sign is 3 steps away, posing a risk of collision. At 2 o'clock, a barrier post is 4 steps away, which could also cause a trip. To safely navigate the area, move slightly to your left and proceed forward, avoiding the central obstacles.\\

Scene Description: \textbf{\{vlm\_output}\}\\
Object Info:\\
\textbf{\{object\_info}\} \\
Guidance: 

\end{tcolorbox}
\caption{The 3-shot prompt designed for SM to generate accessibility-aware guidance generation in \benchmarkname. The inputs of the prompts are \textbf{boldfaced}.}
\label{fig:prompt_apdx_guidedog_sm_few}
\end{figure*}

\section{AI Usage}
We used Gemini to fix typos and polish the written sentences.

\end{document}